\documentclass[sigconf]{acmart}

\AtBeginDocument{%
  \providecommand\BibTeX{{%
    \normalfont B\kern-0.5em{\scshape i\kern-0.25em b}\kern-0.8em\TeX}}}

\setcopyright{acmcopyright}
\copyrightyear{2018}
\acmYear{2018}
\acmDOI{XXXXXXX.XXXXXXX}

\usepackage{url}
\usepackage{multirow}
\usepackage{graphicx}
\usepackage{color,xcolor,colortbl,array}

\acmConference[ ]{ }{ }{ }

\renewcommand\footnotetextcopyrightpermission[1]{}
\begin{document}

\title{DifFSS: Diffusion Model for Few-Shot Semantic Segmentation}


\author{Weimin Tan}
\affiliation{%
  \institution{School of Computer Science, Fudan University}
  \city{Shanghai}
  \country{China}
  }
\email{wmtan@fudan.edu.cn}

\author{Siyuan Chen}
\affiliation{%
  \institution{School of Computer Science, Fudan University}
  \city{Shanghai}
  \country{China}
  }
\email{siyuanchen22@m.fudan.edu.cn}

\author{Bo Yan}
\affiliation{%
  \institution{School of Computer Science, Fudan University}
  \city{Shanghai}
  \country{China}
  }
\email{byan@fudan.edu.cn}
\renewcommand{\shortauthors}{Siyuan, et al.}

\begin{abstract}
Diffusion models have demonstrated excellent performance in image generation. Although various few-shot semantic segmentation (FSS) models with different network structures have been proposed, performance improvement has reached a bottleneck. This paper presents the first work to leverage the diffusion model for FSS task, called DifFSS. DifFSS, a novel FSS paradigm, can further improve the performance of the state-of-the-art FSS models by a large margin without modifying their network structure. Specifically, we utilize the powerful generation ability of diffusion models to generate diverse auxiliary support images by using the semantic mask, scribble or soft HED boundary of the support image as control conditions. This generation process simulates the variety within the class of the query image, such as color, texture variation, lighting, $etc$. As a result, FSS models can refer to more diverse support images, yielding more robust representations, thereby achieving a consistent improvement in segmentation performance. Extensive experiments on three publicly available datasets based on existing advanced FSS models demonstrate the effectiveness of the diffusion model for FSS task. Furthermore, we explore in detail the impact of different input settings of the diffusion model on segmentation performance. Hopefully, this completely new paradigm will bring inspiration to the study of FSS task integrated with AI-generated content. Code is available at \url{https://github.com/TrinitialChan/DifFSS}
\end{abstract}




\maketitle

\section{Introduction}

\textbf{F}ew-shot \textbf{S}emantic \textbf{S}egmentation (\textbf{FSS}) is a difficult task that involves predicting dense masks for new classes with only a limited number of annotations \cite{pl,crnet,fwbnet,pgnet,fss1000}. The key challenge of FSS is to fully utilize the precious information contained in the already scarce support set. Previous works in this field have relied on the idea of prototyping \cite{prototype}, which involves abstracting information from support images into class-wise prototypes through average pooling or clustering, and then matching query features against these prototypes for segmentation label prediction \cite{sgone,fwbnet}. However, this approach can lead to information loss in FSS, given the dense nature of segmentation tasks. Recent works have proposed exploring pixel-wise correlations between query features and foreground support features to avoid this issue, and some have even considered background support features to leverage more information. These state-of-the-art FSS approaches \cite{canet,BAM2022,DCAMA2022,HDMNet2023,HSNet2021} highlight the significance of utilizing the limited information available in the support set. \textit{Are there any new schemes to improve the FSS performance besides this?} One possible solution could be to augment the information available in the support set to meet the requirements of the FSS task.

Diffusion probabilistic model (DPM) is an attractive choice for the above question, as it belongs to a class of deep generative models that have recently emerged as a popular research topic in computer vision \cite{song2020score,song2019generative,ho2020denoising,rombach2022high,saharia2022photorealistic}. The conditional DPMs are capable of generating impressive examples with high levels of detail and diversity. Some models, such as Imagen \cite{saharia2022photorealistic} and Latent Diffusion Models (LDMs) \cite{rombach2022high}, have set new standards in generative modeling. The generated images are of high quality, with very few artifacts, and match well with the given text prompts. The prompts are intentionally chosen to represent unrealistic scenarios that were not seen during training, demonstrating the high generalization ability of diffusion models.

Recently, diffusion models have been used extensively in various generative modeling tasks, such as image super-resolution \cite{rombach2022high}, image inpainting \cite{ho2020denoising,rombach2022high,saharia2022photorealistic}, image generation \cite{rombach2022text}, image-to-image translation \cite{choi2021ilvr}, $etc$. Additionally, the latent representations learned by diffusion models have also shown promise in discriminative tasks, including image classification \cite{zimmermann2021score}, segmentation \cite{baranchuk2021label}, and anomaly detection \cite{wyatt2022anoddpm}. These results indicate that diffusion models have broad applicability and suggest that additional applications are likely to be discovered in the future.

Inspired by this trend, we adapt the diffusion model to the FSS task and experimentally find that the state-of-the-art FSS approaches benefit from this new paradigm significantly and consistently. The effectiveness of diffusion models for FSS tasks can be attributed to two main factors. 1) From a macro perspective, it addresses the imbalance between ``structural'' and ``non-structural'' factors. In existing FSS datasets, it is challenging for semantic objects in different images to have the same pose and structure, with most methods focusing on the ``structural'' dimension. However, ``non-structural'' factors such as texture, color, variety, and light have not been thoroughly explored due to collection deviations in real datasets. The development of generative models enables the full exploitation of such factors. 2) From a micro perspective, rich and diverse auxiliary support images can be created based on the structural information in the support image. This helps the model capture more robust and discriminative category representations, leading to more accurate segmentation of the query image.

In summary, the contributions of this work are the following. 
\begin{itemize}
\item  This is the first work to leverage the diffusion model for few-shot semantic segmentation. Unlike previous FSS methods that focus on designing complex network structures to extract information from the support set, we leverage the diffusion model to generate diverse auxiliary support images that can effectively increase the support set information and enhance the segmentation accuracy of existing FSS models.
\item  Extending the diffusion model to few-shot semantic segmentation presents a promising solution for addressing the imbalance between ``structural'' and ``non-structural'' factors within a class, enabling the extraction of more robust and discriminative category representations and ultimately improving segmentation accuracy. 
\item  We have investigated the effects of varying input configurations of the diffusion model on segmentation performance, providing a thorough analysis of the proposed approach. Our findings may serve as a valuable reference for future studies integrating AI-generated content into few-shot semantic segmentation.
\end{itemize}

\section{Related Work}
This work is related to few-shot segmentation and the diffusion probabilistic model, so in this section, we focus on reviewing the representative works of these two areas.

\subsection{Few-Shot Segmentation}
To obtain the representative vector of each class in FSS task, most works adopt the prototype concept in PN \cite{ppn}, with variations in how the prototype is utilized. For instance, SG-One \cite{sgone} uses the cosine metric to calculate the similarity between the prototype and the query feature, and then merges the similarity map with the query branch through pixel-wise multiplication, which can be thought of as similarity-guided spatial attention. FWB \cite{fwbnet} follows the same process to obtain the similarity map, but replaces the multiplication fusion approach with concatenation. Interestingly, CANet \cite{canet} concatenates the prototype to all spatial locations in the query feature for comprehensive comparisons. The executions of FWB \cite{fwbnet} and CANet \cite{canet} are reversed, with FWB first computing the similarity and then concatenating the feature map, while CANet first concatenates the prototype and then makes comparisons through the convolutional operation. This architecture provides more accurate predictions without other improvement strategies.

Recent state-of-the-art methods, in general, have adopted the dense comparison module from CANet \cite{canet}, such as CyCTR \cite{CyCTR2021}, PMMs \cite{pmm}, PFENet \cite{pfenet}, ASGNet \cite{Li2021AdaptivePL}, BAM \cite{BAM2022}, and HDMNet \cite{HDMNet2023}. To address the problem of incomplete representation caused by using only one foreground prototype, ASGNet adopts the superpixel-guided clustering to obtain multiple ``sub-prototypes''. PAM \cite{cheng2022holistic} is designed to alleviate the problem of overfitting to the targets of base classes. All of these FSS approaches have made significant progress, but are still constrained by the prototype concept framework.

\subsection{Diffusion Probabilistic Model}
The diffusion probabilistic model was first introduced in \cite{sohl2015deep} and has undergone various improvements in both training and sampling methods, such as Denoising Diffusion Probabilistic Model (DDPM) \cite{ho2020denoising}, Denoising Diffusion Implicit Model (DDIM) \cite{nichol2021improved}, and score-based diffusion \cite{zimmermann2021score}. While image diffusion methods can use pixel colors directly as training data, handling high-resolution images requires more computation power, which is often addressed by using pyramid-based or multiple-stage methods \cite{ho2022cascaded} based on the UNet architecture. To reduce the computation power required for training a diffusion model, Latent Diffusion Model (LDM) \cite{wang2018high} was proposed based on the concept of latent image, which was further extended to Stable Diffusion. To support additional input conditions for large pretrained diffusion models, ControlNet \cite{zhang2023adding} was proposed to augment models like Stable Diffusion to enable conditional inputs.

Several works have investigated the application of conditional DPMs in various downstream tasks such as super-resolution, text-to-image synthesis, image inpainting, image segmentation, and instance segmentation \cite{ho2020denoising,rombach2022high,saharia2022photorealistic,rombach2022high,rombach2022text,choi2021ilvr,zimmermann2021score,baranchuk2021label}. However, as far as we know, the proposed DifFSS is the first work to adapt conditional DPMs for the FSS task.

\section{Method}
\subsection{Problem Setup}
Few-shot segmentation aims to segment the query image and predict the semantic mask of the same object category as the support images. We follow the commonly used episodic paradigm \cite{ppn}. Specifically, each episode is composed of a support set $\mathcal{S}$ and a query set $\mathcal{Q}$. For a $K$-shot segmentation task, the support set $\mathcal{S}$ has $K$ image-label pairs $\mathcal{S}=\left\{\left(\mathbf{I}^{\mathrm{s}}_i, \mathbf{M}^{\mathrm{s}}_i\right)\right\}_{i=1}^K \text {, where } \mathbf{I}^{\mathrm{s}}_i \text { and } \mathbf{M}^{\mathrm{s}}_i$ denote the support image and the corresponding annotation mask of the $i$-th data. Similarly, the query set can be denoted by $\mathcal{Q}=\left\{\mathbf{I}^{\mathrm{q}}, \mathbf{M}^{\mathrm{q}}\right\}$, including a query image $\mathbf{I}^{\mathrm{q}}$ and its segmentation annotation $\mathbf{M}^{\mathrm{q}}$. Note that $\mathbf{M}^{\mathrm{q}}$ is only used to evaluate the prediction results $\mathbf{\hat{M}}^{\mathrm{q}}$ of the FSS model during the testing phase.

\begin{figure}[tb]
    \includegraphics[width=8.5 cm]{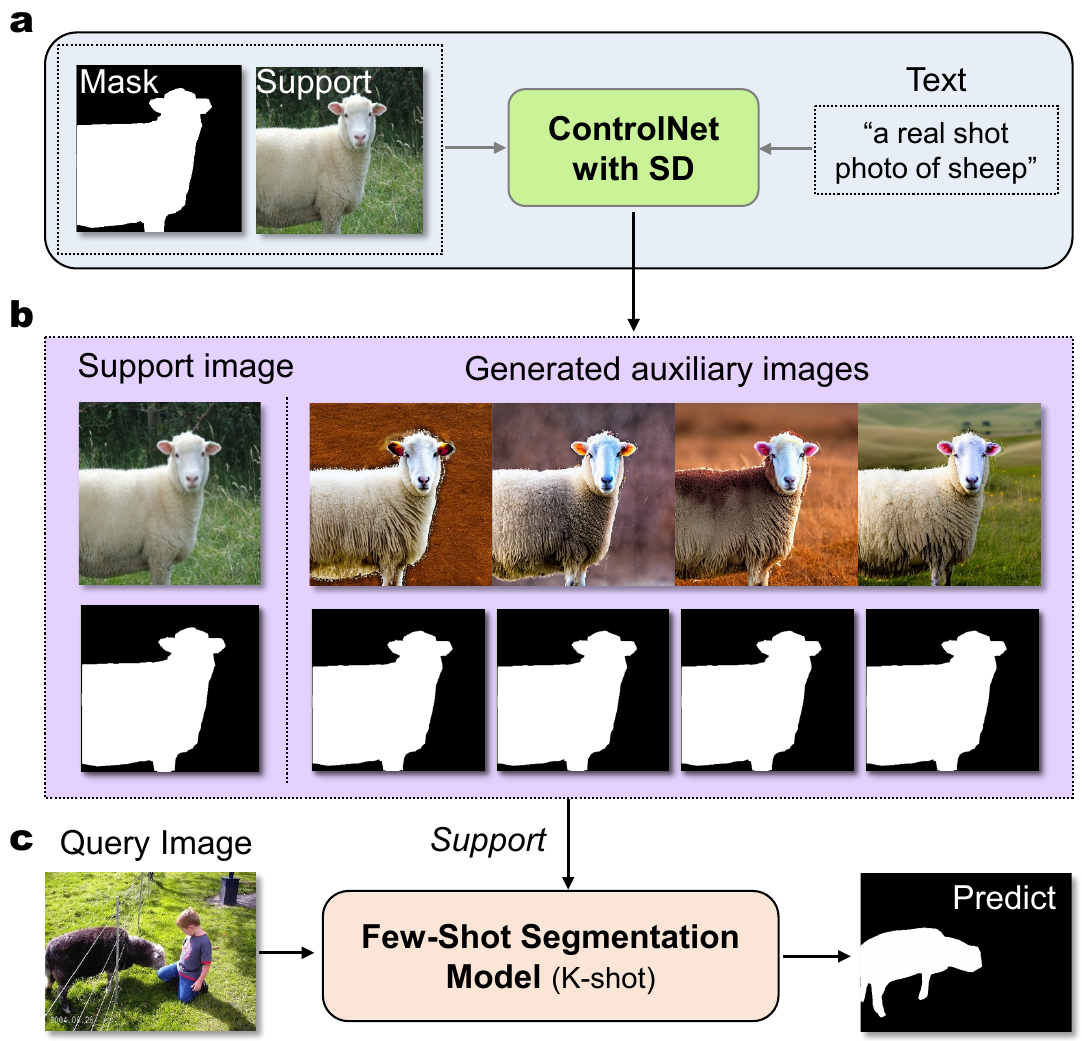}
    \caption{Overview of the diffusion model for few-shot semantic segmentation. It mainly includes three steps. a) The support image, the corresponding segmentation mask, and the prompt text ($e.g.$, “a real shot photo of sheep”, where ``sheep'' is the class name) are inputted into large diffusion models (like ControlNet with Stable Diffusion) to generate auxiliary images that have the same class with the support image. b) The support image and the generated auxiliary images share the same segmentation mask but have quite different appearances and backgrounds. c) These images are used to guide the few-shot segmentation model to segment the query image and predict the semantic mask of the same object category as the support.}
    \label{Fig1_label}
\end{figure}

\begin{figure}[tb]
    \includegraphics[width=8.5 cm]{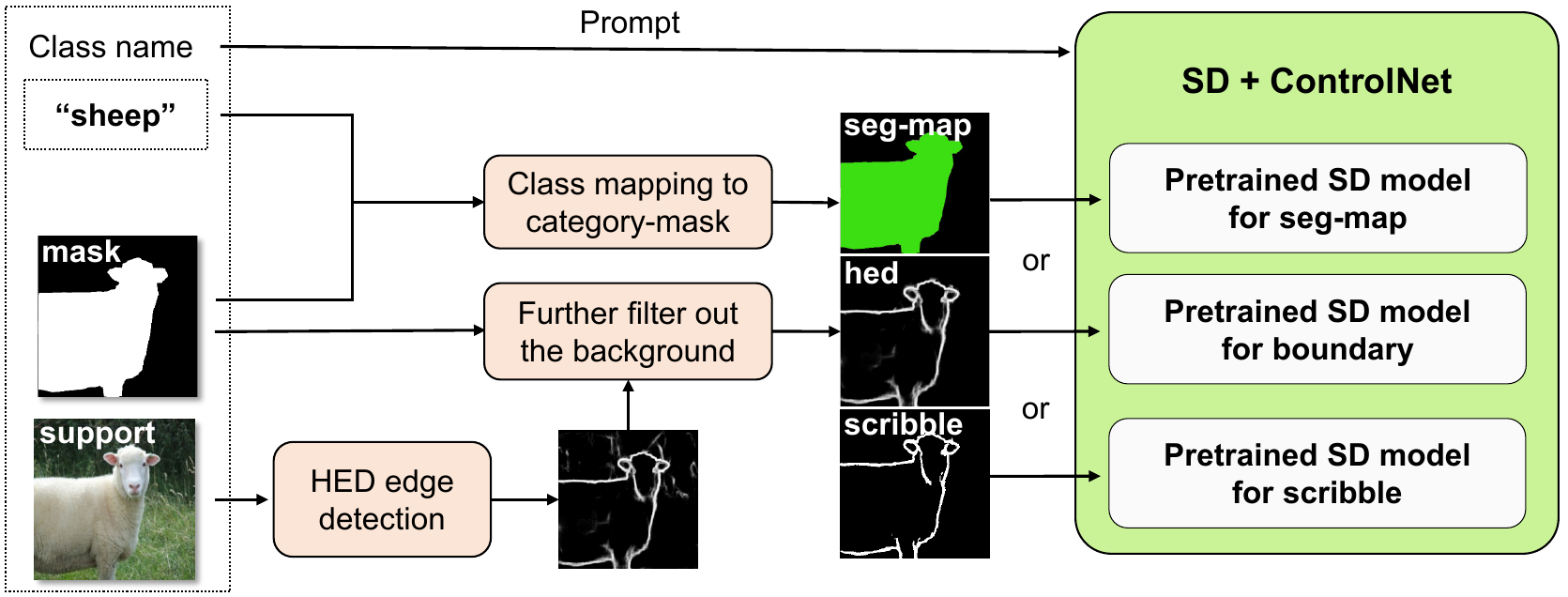}
    \caption{Illustration of the control condition generation for FSS task. In addition to Prompt, one of the segmentation map, boundary map or scribble of the support image is used as the input condition of the diffusion model ($e.g.$, ControlNet with Stable Diffusion).}
    \label{Fig2_label}
\end{figure}

\begin{figure}[tb]
    \includegraphics[width=8.5 cm]{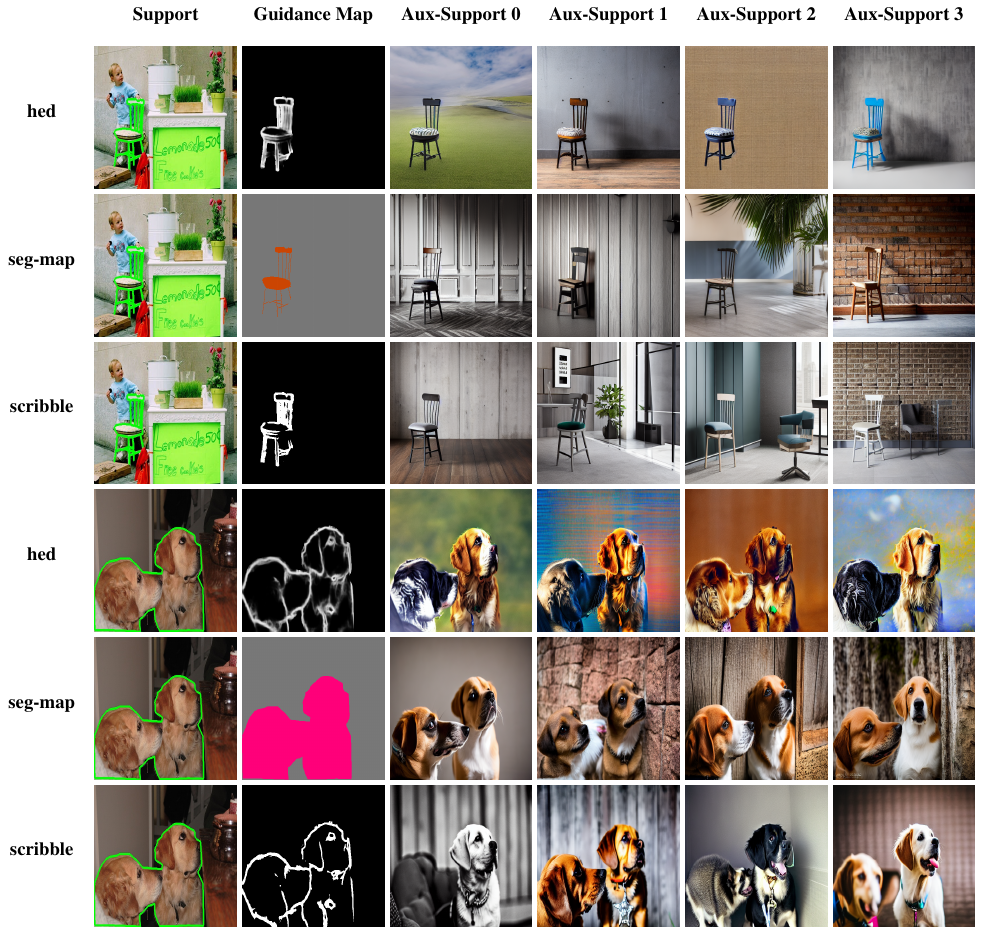}
    \caption{Visual examples of auxiliary images generated under different input control conditions, namely the segmentation map, boundary map, and scribble of the support image. Column two shows the control guidance map.}
    \label{Fig3_label}
\end{figure}

\subsection{Diffusion Model for Few-Shot Segmentation}
\textbf{Motivation and Inspiration}. Given $K$-shot segmentation task, previous FSS methods \cite{BAM2022,pfenet,Li2021AdaptivePL,HDMNet2023} typically use $K$ support images of novel class to activate query features, which is prone to fail to identify the query features due to intra-class diversity, resulting in almost complete failure of segmentation results. For example, ``Bird'' is a huge category, which also contains a lot of sub-categories. For 1-shot segmentation task, only one support image containing a sub-category bird is used to guide the FSS model to segment another sub-category bird in the query image, which is too demanding for segmentation models. Apparently, more support images with large intra-class diversity can significantly alleviate this difficulty. 

To this end, we propose a novel few-shot segmentation paradigm that leverages the powerful capabilities of current image diffusion models to generate intra-class diversity samples. The content generation process is similar to simulating the variety within the class of the query image, such as color, texture variation, light conditions, \textit{etc}. As a result, few-shot segmentation can refer to more diverse support images, yielding more robust representations, thereby achieving a consistent improvement in segmentation performance.

\textbf{Overall Pipeline}. We use the ControlNet with Stable Diffusion \cite{zhang2023adding} as an example to introduce the proposed DifFSS paradigm. The overall pipeline of the DifFSS is shown in Fig. \ref{Fig1_label}. Given the support images and the corresponding annotations of the raw data, we first generate auxiliary support images that have the same class as the support image through the pre-trained ControlNet with Stable Diffusion. This generation process requires the control inputs of the support image, the corresponding segmentation mask, and the prompt text.

\begin{equation}
\mathbf{I}^G_k=\mathbf{G}_\mathcal{F}\left(\mathbf{\Phi}_k\left(\mathbf{I}^{\mathrm{s}}, \mathbf{M}^{\mathrm{s}}\right), Prompt\right)
\end{equation}
where $\mathbf{G}_\mathcal{F}(.)$ denotes the diffusion model. $ \mathbf{\Phi}_k(.)$ denotes the control condition generating function, which performs basic image processing on the input $\mathbf{I}^{\mathrm{s}}$ and $\mathbf{M}^{\mathrm{s}}$, and outputs the control generation conditions like edge maps, segmentation maps, boundaries, $etc$. $Prompt$ is the text in our experiments and is set to the template form ``a real shot photo of \{class name\}''. $\mathbf{I}^G_k$ denotes the generated $ k $-th image. We denote by $\mathbf{I}^G=\left\{ \mathbf{I}^G_1, \mathbf{I}^G_2, \cdots, \mathbf{I}^G_n \right\}$ all $n$ images generated.

Then, the $K$-shot segmentation is formulated as
\begin{equation}
\hat{\mathbf{M}}_q=\mathbf{f}_{seg}\left(\mathbf{I}^q \mid \mathbf{I}^s, \mathbf{M}^s, \mathbf{I}^G\right)
\end{equation}
where $\mathbf{f}_{seg}(.) $ denotes the FSS model. $\mathbf{I}^G$ is the newly added segmentation condition.

Finally, based on the ground-truth label ${\mathbf{M}}_q$ and the prediction $\hat{\mathbf{M}}_q$, we can optimize the FSS model $\mathbf{f}_{seg}(.) $ using the basic cross-entropy loss, with the diffusion model parameters frozen.


\textbf{Control Condition Generation}. Stable Diffusion is a powerful large-scale text-to-image diffusion model, while ControlNet \cite{zhang2023adding} can enhance pretrained image diffusion models with task-specific conditions. To create the control conditions, given the support images and the corresponding labels of the dataset, we first perform HED edge detection \cite{xie2015holistically} on the support images, yielding the edge maps, as shown in Fig. \ref{Fig2_label}. Then, based on the object mask of support images, we can easily filter out the background of edge maps, producing HED boundary maps with clean background. Furthermore, we can obtain the scribble maps by performing binary thresholding ($T=128$, the average intensity for 8-bit image) on the HED boundary maps. Finally, we can obtain the semantic segmentation map by mapping the object class to the support mask. These input conditions will result in different diffusion results because of the difference in their constraint strength, as shown in Fig. \ref{Fig3_label}.

\subsection{Extension to $X$-Shot Setting}
In contrast to previous few-shot semantic segmentation approaches that rely on complex network structures to extract information from support images and annotations to improve segmentation performance, the proposed DifFSS offers a unique advantage in that it can generate numerous auxiliary images based on the given support images and segmentation annotations, enabling an easy extension from a $K$-shot setting to an $X$-shot setting, where $X>K$. For instance, for the 1-shot setting, DifFSS can produce four additional support images, effectively extending the 1-shot setting to the 5-shot setting. In addition, DifFSS can be utilized for zero-shot segmentation since the edge map or boundary map of the query image can act as control conditions for ControlNet with Stable Diffusion. These benefits serve as a good illustration of the numerous and distinct advantages that DifFSS offers to the FSS task.

\subsection{Generation Drift}
We experimentally find that with more generated samples, better FSS performance is usually obtained, but there is a problem that needs to be paid attention to in the extension to $X$-shot setting, $i.e.$, the generation drift problem. This refers to the discrepancy between the position of the semantic object in the generated image and the corresponding object mask in the support image. If the quality of the support image is poor, for instance, when dealing with small objects, occlusions, transparent materials, $etc$., the generation drift becomes more pronounced. As the generated auxiliary images and the original support image share the same object mask, a large drift can result in the use of misleading support images, leading to reduced FSS performance. To further understand the generation drift problem, we will provide a detailed discussion in the experimental section.

\section{Experiment}
To verify the effectiveness of the proposal paradigm, based on existing advanced FSS approaches, we conduct extensive experiments on three publicly available datasets in this section. Furthermore, we discuss in detail the impact of different input settings of the diffusion model on segmentation performance.

\subsection{Datasets and Metric}
\textbf{PASCAL-$5^i$} \cite{oslsm} \textbf{and} \textbf{FSS-1000} \cite{fss1000}. We evaluate the effectiveness of the diffusion model for FSS task on three standard FSS benchmarks: PASCAL-$5^i$, FSS-1000, and COCO-$20^i$. PASCAL-$5^i$ comprises 20 classes that are divided into four folds evenly, $i.e.$, 5 classes per fold. FSS-1000 contains 1,000 classes that are split into 520 training, 240 validation, and 240 testing classes, respectively. 

\textbf{MiniCOCO-$20^i$}. COCO-$20^i$\cite{fwbnet} is a large benchmark created from the COCO dataset. It contains 80 classes that are evenly divided into four folds. We followed the HoughNet method\cite{samet2020houghnet} and reduced the COCO-20 training set to 10\% of the original size. This method ensures that the distribution of the number of categories and the semantic objects of different sizes within a category are consistent with the original training set. In terms of the verification set, we intersect the verification set of COCO2017 (5,000 images) with the COCO-$20^i$ (40,137 images). In addition, to ensure that each category has enough images for 5-shot experiments, we additionally introduced one image from the COCO-$20^i$ validation set. Finally, we constructed a dataset containing 8,200 training and 4,953 validation images, and named it MiniCOCO-$20^i$. Like COCO-$20^i$, it also contains 80 classes.

\textbf{Metric}. For the metric, we adopt mean intersection over union (mIoU) [24] as the main evaluation metric. mIoU is defined as: $\mathrm{mIOU}=1 / C \sum_{i=1}^C \operatorname{IoU}_i$, where $IoU_i$ is the intersection-over-union for class $i$. We follow common practices of previous works for fair comparisons.

\subsection{Implementation Details}
We performed all experiments on the PASCAL-$5^i$ and MiniCOCO-$20^i$ datasets using four NVIDIA GeForce RTX 3090, and all experiments on the FSS-1000 dataset using two NVIDIA GTX 1080Ti. When verifying the effectiveness of our method on these three datasets, we modified the training code of the corresponding models to adapt the proposed DIfFSS framework. Specifically, we initialize the dataloader with a 1-shot configuration and the network model with a 5-shot configuration, so that one support image from the dataset and four auxiliary images from the generative model together with the query image constitute a training sample. For fair comparison, the settings of all hyperparameters and training parameters in the training process follow the settings in the original model. In particular, we regenerated the dataset partition files for BAM  \cite{BAM2022} and HDMNet \cite{HDMNet2023} in the way of PFENet \cite{pfenet} when conducting experiments on MiniCOCO-$20^i$.

\begin{table}[tb]
\caption{Comparison with state-of-the-art methods for 1-shot segmentation on PASCAL-$5^i$ using the mIoU (\%). CyCTR (NeurIPS2021), BAM (CVPR2022), and HDMNet (CVPR2023) are consistently improved after being combined with the diffusion model.}
\label{table_PASCAL}
\resizebox{\columnwidth}{!}{
\begin{tabular}{cllllll}
\toprule
\multicolumn{1}{l}{\multirow{2}{*}{Backbone}} & \multirow{2}{*}{Methods} & \multicolumn{5}{c}{1-shot}                                                    \\ \cline{3-7} 
\multicolumn{1}{l}{}                          &                          & Fold-0        & Fold-1        & Fold-2        & Fold-3        & Mean          \\
\hline 
\multirow{11}{*}{ResNet-50}                   & HSNet \cite{HSNet2021}                    & 64.3          & 70.7          & 60.3          & 60.5          & 64.0            \\
                                              & PFENet \cite{pfenet}                  & 61.7          & 69.5          & 55.4          & 56.3          & 60.8          \\
                                              & SSP   \cite{fan2022self}                   & 60.5          & 67.8          & 66.4          & 51.0            & 61.4          \\
                                              & DCAMA \cite{DCAMA2022}                    & 67.5          & 72.3          & 59.6          & 59.0            & 64.6          \\
                                              & SD-AANet \cite{zhao2023self}                & 60.9          & 70.8          & 58.4          & 57.3          & 61.9          \\
                                              &  CyCTR \cite{CyCTR2021}                    & 65.7          & 71.0          & \textbf{59.5}          & 59.7          & 64.0          \\
  \rowcolor{blue!12}   & \textbf{Ours (CyCTR)}  & \textbf{69.8} & \textbf{71.5} & 59.0 & \textbf{64.4} & \textbf{66.2} \\
                                              &  BAM  \cite{BAM2022}                    & 69.0          & 73.6          & 67.6          & 61.1          & 67.8        \\
 \rowcolor{blue!12}                                             &   \textbf{Ours (BAM)}       & \textbf{69.9} & \textbf{74.6} & \textbf{68.2} & \textbf{64.4} & \textbf{69.3} \\
                                              & HDMNet \cite{HDMNet2023}                  & 71.0          & 75.4          & 68.9          & 62.1          & 69.4          \\
 \rowcolor{blue!12}    &   \textbf{Ours (HDMNet)} & \textbf{71.3} & \textbf{75.5} & \textbf{69.7} & \textbf{64.3} & \textbf{70.2} \\
 \bottomrule
\end{tabular}
}
\end{table}

\subsection{Comparison with State-of-the-Art Methods}
In Tables 1-3, we demonstrate the comparison with state-of-the-art methods for 1-shot segmentation on PASCAL-$5^i$, FSS-1000, and MiniCOCO-$20^i$ using the mIoU (\%) metric. To verify the generality, several state-of-the-art FSS approaches are combined with the diffusion model. The experimental results demonstrate that all of them gain significant improvement from the diffusion model, and they achieve new state-of-the-art performance on PASCAL-$5^i$, FSS-1000, and MiniCOCO-$20^i$ datasets. On the challenging PASCAL-$5^i$ dataset, consistent and significant improvement is achieved for three FSS approaches (CyCTR, BAM, and HDMNet). Especially on FSS-1000 and MiniCOCO-$20^i$, the FSS methods with the diffusion model outperform the prior arts by a large margin, achieving 1.3\% and 2.4\% of mIoU improvements over the SOTA methods. Note that the control condition uses the best segmentation performance among HED boundary, segmentation, or scribble. Table \ref{table_Ablation1} explores the subtle differences between them in detail.

\begin{figure*}[t]
    \includegraphics[width=16.2 cm]{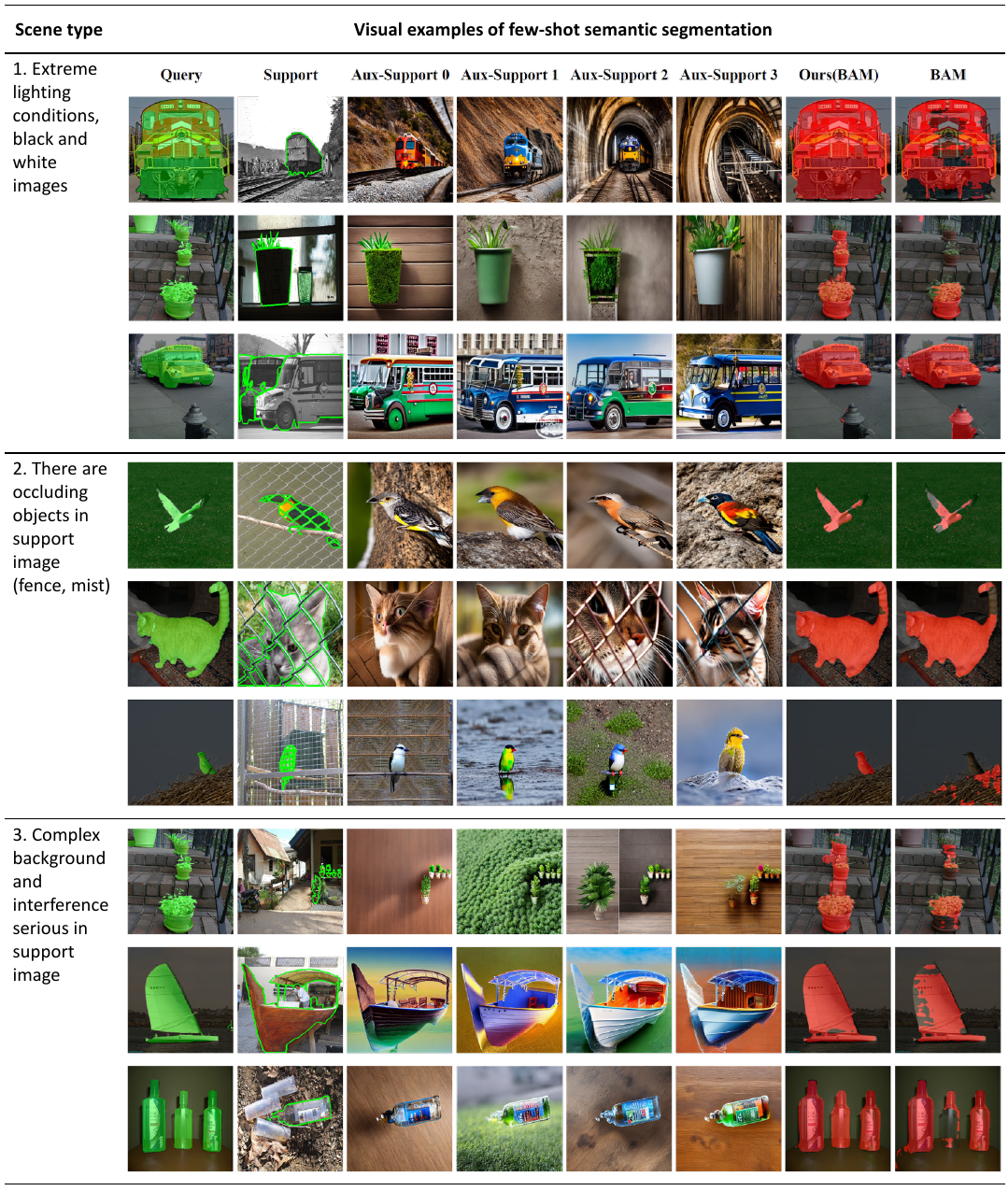}
    \caption{Qualitative segmentation results before and after BAM method \cite{BAM2022} augmented with diffusion models on \textbf{PASCAL-$5^i$}.}
    \label{Fig9_label}
\end{figure*}

\begin{figure*}[t]
    \includegraphics[width=16.2 cm]{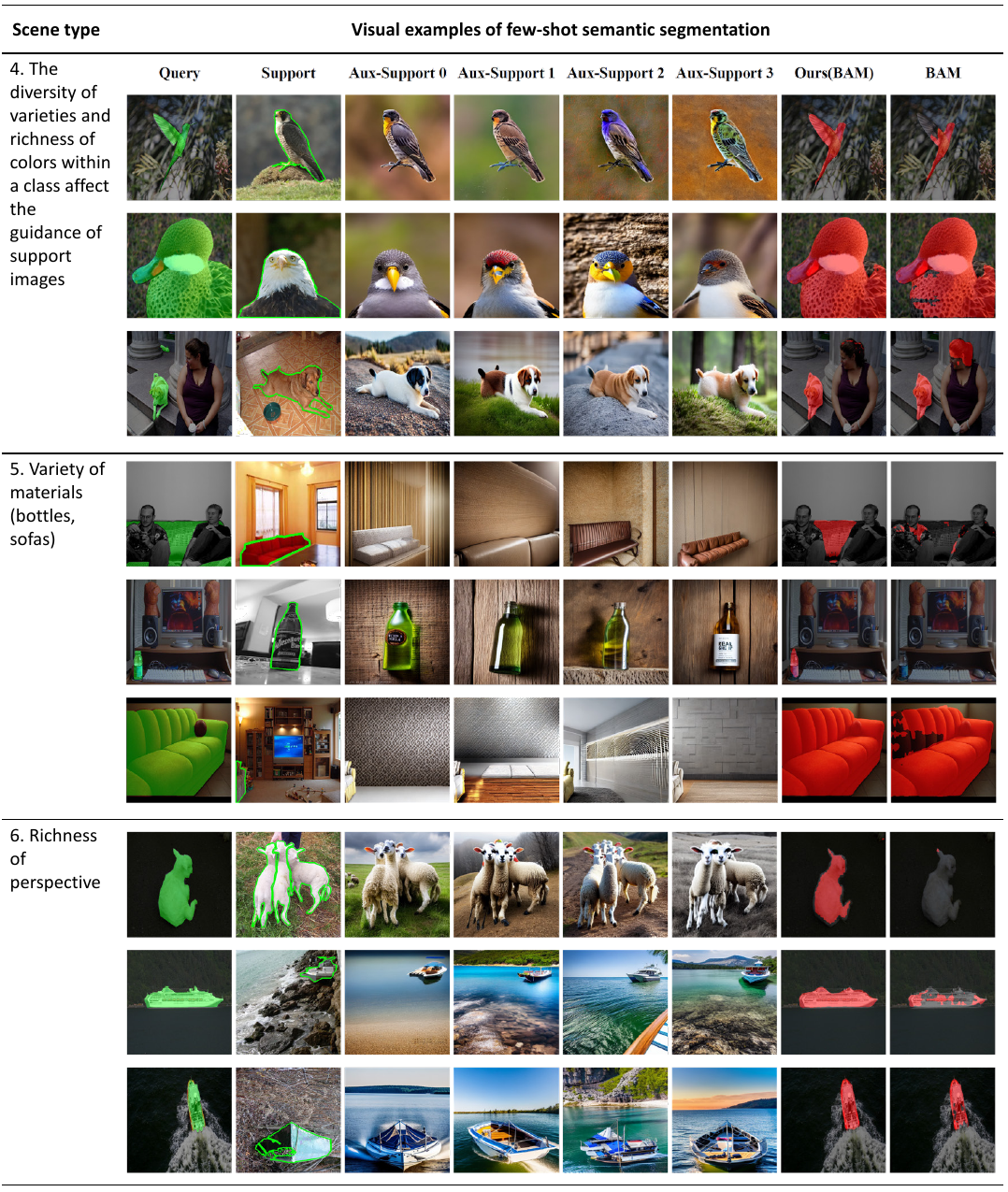}
    \caption{Qualitative segmentation results before and after BAM method \cite{BAM2022} augmented with diffusion models on \textbf{PASCAL-$5^i$}.}
    \label{Fig10_label}
\end{figure*}


\begin{figure*}[t]
    \includegraphics[width=17.5 cm]{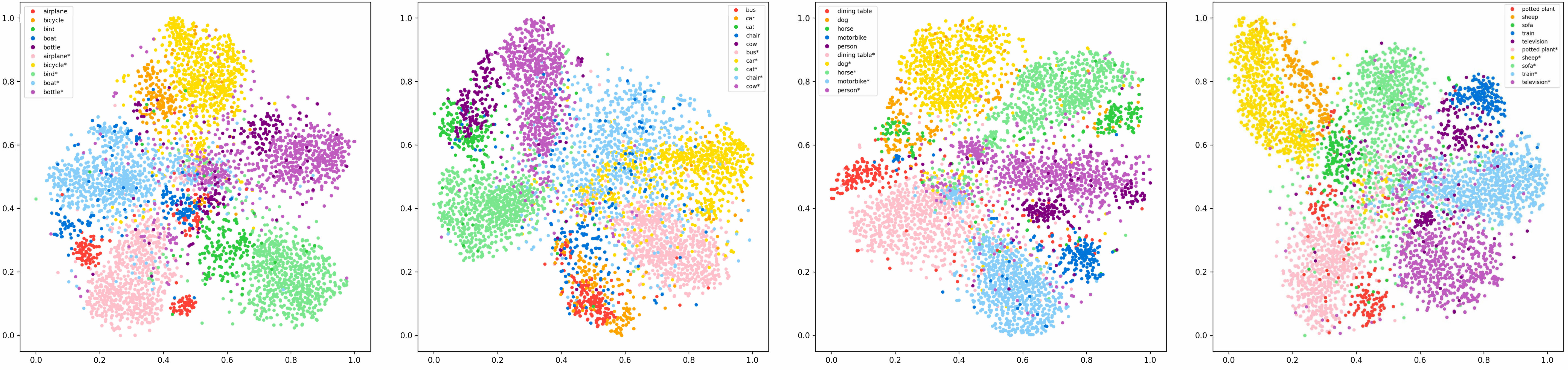}
    \caption{Comparison of prototype distributions between the support images from the raw dataset (dark-colored points) and the images generated by the diffusion model (light-colored points) on PASCAL-$5^i$. Four subgraphs from left to right are the prototype distribution visualization of 20 categories of PASCAL-$5^i$ divided into four groups, each with five categories.}
    \label{Fig14_label}
\end{figure*}

\begin{table}[tb]
\caption{Performance on FSS-1000 using the mIoU (\%). HSNet (ICCV2021), and DCAMA (ECCV2022) are consistently improved after being combined with the diffusion model.}
\label{table_FS1000}
\begin{tabular}{lll}
\toprule
Backbone                   & Methods               & mIoU (1-shot)  \\
\midrule 
\multirow{4}{*}{ResNet-50} & HSNet \cite{HSNet2021}                 & 85.5          \\
                       &  \textbf{Ours (HSNet)} & \textbf{86.2} \\
                           & DCAMA \cite{DCAMA2022}                 & 88.2          \\
                      & \textbf{Ours (DCAMA)} & \textbf{88.4} \\ \bottomrule
\end{tabular}
\end{table}

\begin{table}[tb]
\centering
\caption{Performance on MiniCOCO-$20^i$ using mIoU (\%). BAM (CVPR2022) and HDMNet (CVPR2023) are consistently improved after being combined with the diffusion model.}
\label{table_minicoco$20^i$}
\resizebox{\columnwidth}{!}{
\begin{tabular}{lllllll}
\toprule
\multirow{2}{*}{Backbone}  & \multirow{2}{*}{Methods} & \multicolumn{5}{c}{1-shot}                                                    \\ \cline{3-7} 
 &               & Fold-0 & Fold-1 & Fold-2 & Fold-3 & \textbf{Mean} \\ \hline
\multirow{4}{*}{ResNet-50} & BAM \cite{BAM2022}                      & 37.6          & 47.8          & 36.9          & 41.0          & 40.8          \\
                           &  \textbf{Ours (BAM)}      & 40.5 & 48.7 & 43.0 & 42.2 & \textbf{43.6} \\
 & HDMNet \cite{HDMNet2023}        & 43.0      & 51.2      & 44.7     & 46.4      & 46.3    \\
 &  \textbf{Ours (HDMNet)} & 42.8     & 50.9     & 45.0     & 48.1    & \textbf{46.7}   \\ \bottomrule
\end{tabular}
}
\end{table}

\begin{table}[tb]
\caption{Comparison of different control conditions for BAM \cite{BAM2022} and HDMNet\cite{HDMNet2023}) on PASCAL-$5^i$. The value preceding ``/'' in ``Gain rate'' column represents the mIoU gain achieved by utilizing the diffusion model (one support image and four auxiliary images) in comparison to the 1-shot setting, whereas the value succeeding ``/'' represents the mIoU gain obtained by using the 5-shot setting as opposed to the 1-shot setting.}
\label{table_Ablation1}
\resizebox{\columnwidth}{!}{
\begin{tabular}{llcccl}
\toprule
\multirow{2}{*}{Guide Method} & \multirow{2}{*}{Model} & \multicolumn{3}{c}{mean mIoU} & \multirow{2}{*}{Gain rate} \\ \cline{3-5}
         &        & 1-shot                & \textbf{1-shot (with 4 aux.)} & 5-shot                &          \\ \hline
seg-map  & BAM    & \multirow{3}{*}{67.8} & \textbf{69.2}              & \multirow{3}{*}{70.9} & \textcolor{red}{+1.4}/\textcolor{blue}{3.1} \\
hed      & BAM    &                       & \textbf{69.2}              &                       & \textcolor{red}{+1.4}/\textcolor{blue}{3.1} \\
scribble & BAM    &                       & \textbf{69.3}              &                       & \textcolor{red}{+1.5}/\textcolor{blue}{3.1} \\ \hline
seg-map  & HDMNet & \multirow{3}{*}{69.4} & \textbf{70.0}              & \multirow{3}{*}{71.8} & \textcolor{red}{+0.6}/\textcolor{blue}{2.4} \\
hed      & HDMNet &                       & \textbf{69.8}              &                       & \textcolor{red}{+0.4}/\textcolor{blue}{2.4} \\
scribble & HDMNet &                       & \textbf{70.2}              &                       & \textcolor{red}{+0.8}/\textcolor{blue}{2.4} \\ \bottomrule
\end{tabular}%
}
\end{table}

\subsection{Qualitative Results}
We present a lot of qualitative segmentation results before and after using our DifFSS in this section according to the scene type. Based on our extensive observations in experiments, the scene type is roughly divided into eight categories, as shown in Fig. \ref{Fig9_label} and Fig. \ref{Fig10_label}. Due to limited space, please see the \textit{supplementary file} for more examples of scene types. These intuitive examples demonstrate the significant improvement brought by the diffusion model to FSS performance.

\begin{figure*}[tb]
    \includegraphics[width=17.5 cm]{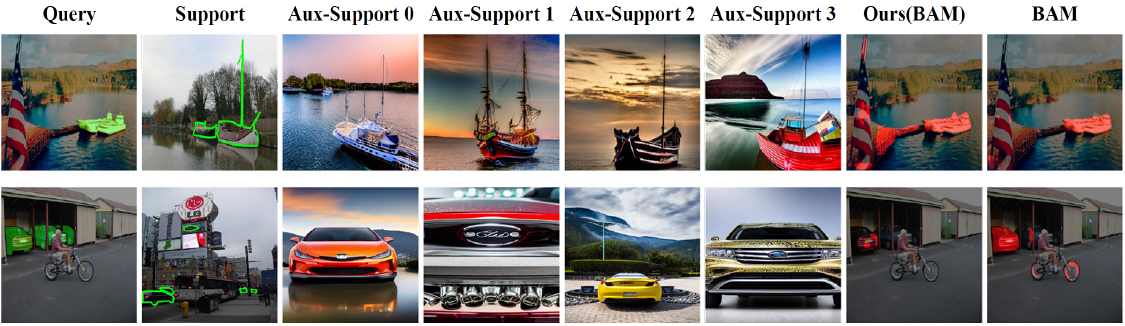}
    \caption{Visualization of generation drift. Small and scattered objects in the support image are likely to cause generation drift.}
    \label{Fig12_label}
\end{figure*}

\subsection{Ablation Study}
We report the ablation study in this section to investigate the impact of each control guidance, the performance of extending to $X$-Shot Setting, and the consistency of prototype distribution. 

\textbf{Guidance Ablation}. As mentioned in the Methods section, different control guidance typically generates different images from the ``non-structural'' dimension ($e.g.$, color, pose, texture, lighting, $etc$.). We are interested in how much each guidance contributes to the FSS task and whether there are large differences. To address this doubt, we further experiment with two SOTA methods (BAM \cite{BAM2022} and HDMNet\cite{HDMNet2023}) on PASCAL-$5^i$ with auxiliary images of different input conditions. Table \ref{table_Ablation1} reports the guidance ablation results. Firstly, all guidance types consistently improve the mIoU of BAM and HDMNet methods under the 1-shot setting. In addition, compared with the true 5-shot method, $i.e.$, the five support images are from the original dataset, there is still a lot of room for improvement in the proposed pseudo-5-shot method (one support image and four generated auxiliary images). Therefore, developing generative models and generative conditions suitable for FSS tasks is well worth further exploration in the future.

\textbf{$\mathbf{X}$-Shot Setting Ablation}. We initialize the dataloader and network model with the 1-shot configuration. The model input is one support image from the raw dataset and $n$ auxiliary images from the generated model, which together with the query image will constitute $n+1$ training samples for model training. HSNet method \cite{HSNet2021} shows impressive segmentation performance and thus is selected as the test baseline. As shown in Table \ref{table_Ablation2}, the FSS performance generally increases with the number of auxiliary images. Compared with the none auxiliary, the mIoU of HSNet with four auxiliary images increases from 85.19\% to 86.20\%. The result well demonstrates the effectiveness of generated auxiliary images for FSS task.

\textbf{Prototype Distribution Ablation}. To further investigate why the diffusion model can enhance the segmentation performance in the FSS task, it is crucial to verify whether the semantic prototype distribution of the image generated by the diffusion model extracted in the FSS model is consistent with that of the original support image. To this end, we visualize their prototypical distributions for all 20 categories on the PASCAL-$5^i$ dataset. Specifically, we extract the features of the FSS model based on ResNet50 and take the feature maps (1024$\times$H$\times$W) output by Block-3 of ResNet50 for each image, followed by obtaining the prototype (1024$\times$1) after masked average pooling and performing T-SNE dimensionality reduction after L2 normalization. 

Comparison of prototype distributions between the raw support images and the generated images on PASCAL-$5^i$ is shown in Fig. \ref{Fig14_label}. The dark-colored points represent the original support image, and the corresponding light-colored points represent the generated image of the diffusion model (using scribble as the control condition). The visualization results indicate that most of the light-colored points are clustered near the centroid of the dark-colored points, and the clustering effect of the light-colored points is obvious. This demonstrates that the generated images express rich intra-class diversity while maintaining semantic consistency with the original support image. Moreover, the significant clustering effect implies that the generated images can provide robust semantic representations. Therefore, the auxiliary images generated by the diffusion model can help FSS models capture more robust and discriminative category representations, leading to more accurate segmentation of the query image.


\begin{table}[tb]
\centering
\caption{Ablation study of the number of generated auxiliary images for HSNet (ICCV2021) on FSS-1000. Note that the mIoU reported in HSNet paper \cite{HSNet2021} with Resnet50 backbone is 85.50\%. Our replicated result achieved 85.19\% after 700 epochs of training, which is the best result obtained.}
\label{table_Ablation2}
\begin{tabular}{llllll}
\toprule
\#Aux.       & 0     & 1     & 2     & 3     & \textbf{4}     \\ \hline
mIoU (\%) & 85.19 & 85.23 & 86.16 & 85.91 & \textbf{86.20}   \\ \bottomrule
\end{tabular}%
\end{table}

\begin{table}[tb]
\caption{Quantification of generation drift using the pre-trained UPerNet method to segment the raw support images and the generated images on PASCAL-$5^i$.}
\label{table_GDrift}
\begin{tabular}{lllll}
\toprule
Guide Method & Seg-map & Hed   & Scribble & Original \\ \hline
mIoU(\%)     & 51.34   & \textbf{56.39}$^*$ & 51.56    & \textbf{71.41}   \\ \bottomrule
\end{tabular}
\end{table}

\subsection{Discussion}
\textbf{Generation Drift}. To demonstrate the phenomenon of generation drift, we experiment with the pre-trained UPerNet method \cite{xiao2018unified} on PASCAL-$5^i$ to segment the raw support images and the generated images. Table \ref{table_GDrift} reports the difference between the object positions of the raw support image and the generated images. For the same segmentation method UPerNet \cite{xiao2018unified}, the segmentation result of the raw support image has the highest overlap with its corresponding annotated mask. Among the three control conditions, the image generated based on the Hed boundary has a mIoU of 56.39\% and the generated drift is relatively minimal. Visual examples of generation drift are shown in Fig. \ref{Fig12_label}. The position of the object in the generated image exhibits a large offset from the labeling of the support image. Thus, it is crucial to address the issue of generation drift in the image diffusion for FSS task.

\textbf{Sensitivity to Support Quality}. The term ``support quality'' in this context denotes the level of information and complexity inherent in the support image that FSS models utilize to accurately segment the query image. We experimentally find that the quality of support image has a significant impact on the effectiveness of the diffusion model for FSS task. In fact, generation drift is a good example because of the small and scattered objects in the support image. In practical applications, it is advisable to select high-quality support images to avoid such issues.

\section{Conclusion}
This paper has presented a novel few-shot segmentation paradigm that leverages the diffusion model for FSS task. To verify its effectiveness, based on existing advanced FSS models, extensive experiments on three publicly available datasets demonstrated a consistent and significant performance improvement. Furthermore, we have discussed in detail the impact of different input settings on the diffusion model for segmentation performance. All these experimental results suggest that the advancement of AI-generated content technologies, such as diffusion models, has the potential to introduce novel concepts to tasks such as few-shot semantic segmentation, and may facilitate breakthroughs in performance bottlenecks. We believe that this is a new and exciting direction to explore for FSS task.

\bibliographystyle{ACM-Reference-Format}
\bibliography{main}



\end{document}